\documentclass{article}
\usepackage{graphicx}
\usepackage{amsmath}
\usepackage{amssymb} 
\usepackage{subfig}
\usepackage{enumitem}

\PassOptionsToPackage{numbers,sort&compress}{natbib}
\usepackage[preprint]{neurips_2026}
\usepackage{graphicx}
\usepackage{tikz}
\usetikzlibrary{arrows.meta,positioning,fit,calc,backgrounds}
\usepackage{xcolor}

\usepackage[utf8]{inputenc} 
\usepackage[T1]{fontenc}    
\usepackage{hyperref}       
\usepackage{url}            
\usepackage{booktabs}       
\usepackage{amsfonts}       
\usepackage{nicefrac}       
\usepackage{microtype}      
\usepackage{xcolor}         
\usepackage{graphicx}
\usepackage{amssymb}
\usepackage{pifont}

\newcommand{\cmark}{\ding{51}}
\usepackage[table]{xcolor}
\usepackage{multirow}
\usepackage{booktabs}
\usepackage{float}
\usepackage{needspace}
\usepackage{placeins}
\usepackage{etoolbox}

\AtBeginEnvironment{table}{\small}
\AtBeginEnvironment{table*}{\small}

\title{CoWorld-VLA: Thinking in a Multi-Expert World Model for Autonomous Driving}

\author{
Minqing Huang$^{1\ast}$
\quad
Yujiao Xiang$^{1,2\ast}$
\quad
Zihan Liang$^{1,3\ast}$
\quad
Jiajie Huang$^{1,4\ast}$
\quad
Jingqi Wang$^{1\ast\dagger}$
\quad
\\
\textbf{
Yuheng Zhou$^{1,5}$
\quad
Zhi Xu$^{1}$
\quad
Feiyang Tan$^{1}$
\quad
Hangning Zhou$^{1}$
\quad
Mu Yang$^{1}$
\quad
Gong Chen$^{1,6}$
\quad
}
\\
{\small
$^{1}$ Afari Intelligent Drive
\quad
$^{2}$ University of Electronic Science and Technology of China}
\\
\quad
{\small
$^{3}$ Shanghai Jiao Tong University
\quad
$^{4}$ Beijing University Of Posts and Telecommunications}
\\
\quad
{\small
$^{5}$ Southeast University
\quad
$^{6}$ Tianjin University}
\\
{\small
$^{\ast}$ The authors contributed equally and are listed in no particular order.}
\quad
\\
{\small
$^{\dagger}$ Corresponding author: 
jingqi.wang1314@gmail.com
}
}

\begin{document}

\maketitle

\begin{abstract}
Vision-Language-Action (VLA) models have emerged as a promising paradigm for end-to-end autonomous driving. However, existing reasoning mechanisms still struggle to provide planning-oriented intermediate representations: textual Chain-of-Thought (CoT) fails to preserve continuous spatiotemporal structure, while latent world reasoning remains difficult to use as a direct condition for action generation. In this paper, we propose CoWorld-VLA, a multi-expert world reasoning framework for autonomous driving, where world representations serve as explicit conditions to guide action planning. CoWorld-VLA extracts complementary world information through multi-source supervision and encodes it into expert tokens within the VLA, thereby providing planner-accessible conditioning signals. Specifically, we construct four types of tokens: semantic interaction, geometric structure, dynamic evolution, and ego trajectory tokens, which respectively model interaction intent, spatial structure, future temporal dynamics, and behavioral goals. During action generation, CoWorld-VLA employs a diffusion-based hierarchical multi-expert fusion planner, which is coupled with scene context throughout the joint denoising process to generate continuous ego trajectories. Experiments on NAVSIM v1 demonstrate future-scene modeling capability and strong planning performance, including collision avoidance and trajectory accuracy. Ablation studies further validate the complementarity of expert tokens and their effectiveness as planning conditions for action generation. Code will be available at \url{https://github.com/AFARI-Research/CoWorld-VLA}.
\end{abstract}

\section{Introduction}

\begin{figure*}[t]
    \centering
    \includegraphics[width=\textwidth]{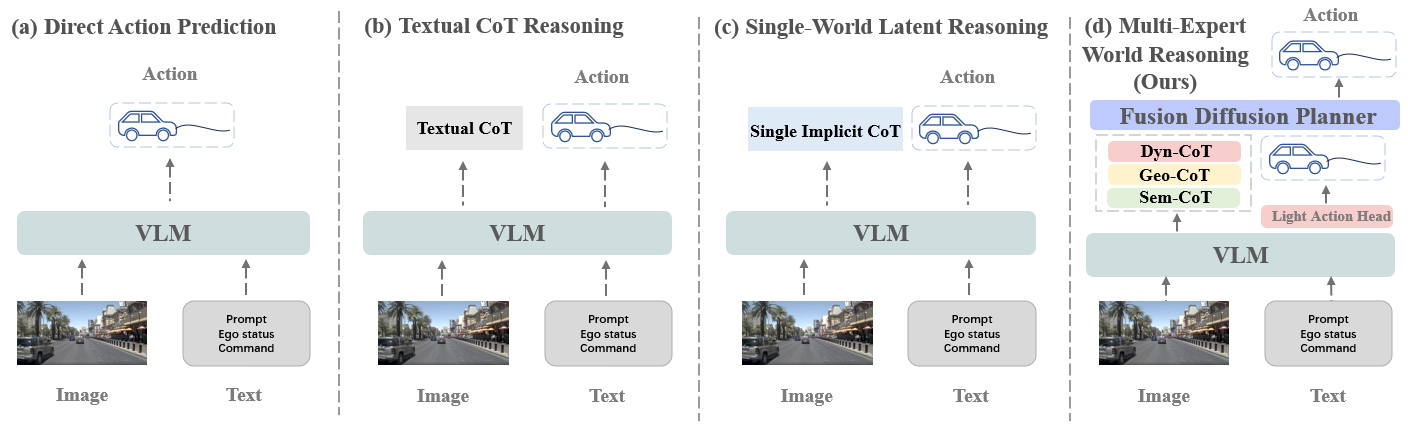}
    \caption{
    \textbf{Comparison of reasoning paradigms for VLA-based autonomous driving.}
    (a) Direct action prediction maps multimodal inputs to actions without intermediate reasoning.
    (b) Textual CoT introduces language-based reasoning but may lose continuous spatio-temporal details.
    (c) Single-world latent reasoning relies on one implicit world representation, which may be incomplete or weakly coupled with actions.
    (d) CoWorld-VLA performs multi-expert world reasoning by organizing Latent CoT experts and using a fusion diffusion planner for trajectory generation.
    }
    \label{fig:introduction}
    \vspace{-10pt}
\end{figure*}

In recent years, foundation models such as Large Language Models (LLMs) and Vision-Language Models (VLMs) have shown strong multimodal understanding and cross-modal generalization capabilities~\cite{12,45,46}, leading to their increasing adoption in robotics and autonomous driving~\cite{47,48}. VLA models have thus emerged as an important paradigm for end-to-end autonomous driving, mapping multimodal observations and language instructions to vehicle actions~\cite{35,60,61,62,63,64,65}. However, autonomous driving requires reasoning about traffic participants, road geometry, future scene evolution, and ego objectives~\cite{6,7,73,74,75}. Existing VLA frameworks often lack explicit intermediate reasoning states, forcing the model to jointly perform scene understanding, future prediction, and trajectory planning within a single action-generation process, which limits performance in complex driving scenarios~\cite{34,43,77}. Figure~\ref{fig:introduction}(a) illustrates existing VLA paradigms.

Recent studies introduce Chain-of-Thought (CoT) into autonomous driving VLA models to improve reasoning in complex scenarios~\cite{13,19,20}. However, most methods rely on explicit natural-language reasoning, which introduces inference overhead and struggles to preserve continuous spatial and motion information for control~\cite{21,22,23,24}. This motivates intermediate representations that preserve spatiotemporal structure and can directly condition trajectory planning.

World models offer a promising alternative by modeling future environment evolution in latent space~\cite{25,27}. They have been widely used for scene generation, dynamic prediction, and planning assistance in autonomous driving~\cite{31,38,41,42}. However, planning depends on multiple forms of world knowledge, including semantic interactions, 3D structure, and dynamic evolution, which are difficult to capture with a single representation~\cite{26,27,29,31,38,41,49,55,56,57}.
 Moreover, predicted world representations are usually used only as auxiliary supervision rather than explicit conditions for trajectory generation during inference~\cite{39,68,82,85}, limiting their effectiveness for planning.

These limitations point to a central bottleneck: current VLA systems lack a mechanism for turning complementary world knowledge into planning-oriented latent states. To address this, we propose CoWorld-VLA, a multi-expert world reasoning framework for autonomous driving (Figure~\ref{fig:introduction}(d)).
 CoWorld-VLA introduces four specialized expert tokens in the VLM latent space to form a planning-oriented Latent CoT: semantic interaction token for high-level intentions and interactions, guided by JEPA-style representations~\cite{56,26}, the geometric structure token for road layouts and spatial constraints, using VGGT features~\cite{55}, dynamic evolution token for future scene evolution, supervised by the generative world model Wan~\cite{11}, and ego-trajectory tokens that connect world reasoning with behavioral objectives through trajectory-level supervision. Together, these tokens condense complementary world knowledge into intermediate reasoning states, alleviating the incompleteness of a single representation.
To make these expert tokens directly support action generation, we further design a diffusion-based hierarchical multi-expert fusion planner. It converts expert tokens into trajectory-generation conditions and progressively produces continuous ego trajectories through denoising. In this way, world knowledge from different sources participates in inference-time planning rather than remaining only an auxiliary training signal.

In summary, our main contributions are as follows:
\begin{itemize}[leftmargin=3em]
\item We propose \textbf{CoWorld-VLA}, a unified multi-expert latent world reasoning framework. CoWorld-VLA formulates intermediate reasoning as a multi-expert Latent CoT in the VLM latent space, modeling semantic interaction, 3D geometry, dynamic evolution, and ego trajectory through expert tokens.

\item We introduce a diffusion-based hierarchical multi-expert fusion planner to bridge the gap between world modeling and action generation. The planner integrates expert tokens with scene context during joint denoising for continuous trajectory generation.

\item Extensive experiments on trajectory planning, future scene generation, and ablation studies demonstrate the effectiveness of CoWorld-VLA. Our method demonstrates future-generation capability and strong planning performance on NAVSIM v1, while ablations validate the complementary effects of multi-expert tokens and Latent CoT.
\end{itemize}

\section{Related work}
\label{sec:related_Work}

\paragraph{Vision-language models for autonomous driving.}

Early VLA studies formulate autonomous driving as language generation. Methods such as DriveGPT4~\cite{13} and DriveLM~\cite{59} leverage LLMs and VQA for driving decision-making, but remain limited by text-based interfaces and cannot directly generate executable trajectories~\cite{14}. Subsequent approaches, including DriveVLM~\cite{61}, Senna~\cite{62}, DiffVLA~\cite{63}, and VLP~\cite{64}, decouple high-level reasoning from low-level control, while methods such as GPT-Driver~\cite{15}, EMMA~\cite{16}, Orion~\cite{65}, and OmniDrive~\cite{66} reformulate trajectory prediction as textual reasoning or generation. However, recent studies show that lengthy textual reasoning may increase inference latency and weaken critical visual information~\cite{2,17,18,19,20}. To address these issues, recent work explores latent reasoning in continuous latent spaces~\cite{21,22,23,24}. Representative methods include DriveMoE~\cite{67}, ReCogDrive~\cite{2}, and LaST-VLA~\cite{luo2026lastvla}. Nevertheless, existing latent reasoning approaches still lack sufficient physical and semantic constraints for structured planning representations.

\paragraph{World models for autonomous driving.}

World models are widely used in autonomous driving to capture spatio-temporal dynamics and predict future scene evolution~\cite{37,38,39,40}. Early works mainly focus on future scene generation, including videos, point clouds, and multimodal signals~\cite{31,68,69,70,71}. More recent approaches jointly model environment evolution and driving policies~\cite{5,25,26,27,28,29,30,31,32,33,34,35,36,41,42}. However, future scene prediction and trajectory planning are often handled by separate branches, limiting the planner’s ability to exploit learned scene-dynamics representations. To address this issue, methods such as Uni-World VLA~\cite{4} and DriveLaW~\cite{43} attempt to unify environment modeling and planning. Nevertheless, RGB-only world models still lack sufficient structural and physical understanding for robust decision-making~\cite{49,50,51}. Recent advances in geometric structure modeling~\cite{52,53,54,55}, JEPA-based predictive learning~\cite{56,57,26}, and video generation models such as Wan~\cite{11} provide complementary geometric, dynamic, and visual priors, but these knowledge sources are rarely aligned within a unified latent reasoning framework for autonomous driving.
\section{Methodology}
\label{sec:methodology}
\begin{figure}[t]
    \centering
    \includegraphics[width=\textwidth]{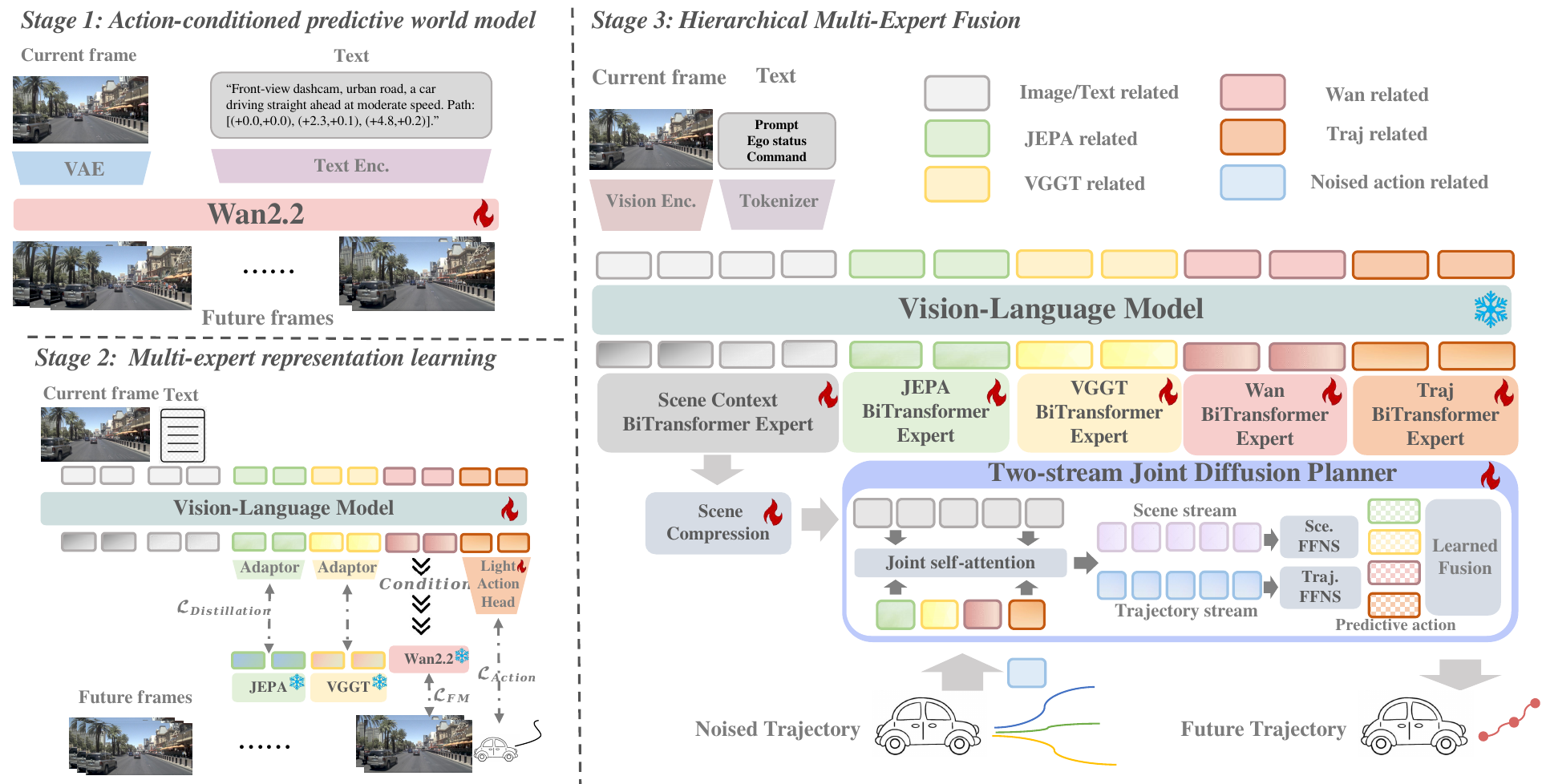}
    \caption{
    \textbf{Overview of CoWorld-VLA}.
    CoWorld-VLA follows a three-stage training pipeline: video-generator pre-training, multi-expert world-representation learning, and diffusion-based trajectory planning. It first learns future scene evolution from visual and textual conditions, then aligns VLM hidden states with semantic, geometric, visual-dynamic, and trajectory experts, and finally fuses these expert representations to generate world-consistent ego trajectories.
    }
    \label{fig:overview}
\end{figure}

Figure~\ref{fig:overview} provides an overview of the proposed CoWorld-VLA framework. We first introduce the necessary preliminaries in Section 3.1. Section~3.2 presents the action-conditioned predictive world model for learning future scene dynamics. Section~3.3 describes multi-expert representation learning, which aligns VLM hidden states with semantic, geometric, visual-dynamic, and trajectory priors. Section~3.4 introduces the hierarchical multi-expert fusion planner for diffusion-based trajectory generation.

\subsection{Preliminaries}

We formulate end-to-end autonomous driving as a conditional action generation problem, where the model predicts the future ego trajectory \(\mathbf{A}_{t+1:t+T}\) over a planning horizon of \(T\) steps.

\paragraph{Input representation.}
At each time step \(t\), the model receives a front-view image \(o_t \in \mathbb{R}^{3 \times H \times W}\) and conditioning information
\(
c_t = \{\kappa_t, \{(\bar{x}_i,\bar{y}_i,\bar{\psi}_i)\}_{i=t-n_h+1}^{t}, v_t, a_t\},
\)
where \(\kappa_t\) is the navigation instruction, \(\{(\bar{x}_i,\bar{y}_i,\bar{\psi}_i)\}\) is the ego pose history, and \(v_t, a_t \in \mathbb{R}^2\) are ego velocity and acceleration.

\paragraph{Standard VLA formulation.}
Existing VLA methods map multimodal inputs directly to future actions~\cite{35,48,60,61,62,63,64,65}:
\begin{equation}
    p_\theta(\mathbf{A}_{t+1:t+T} \mid o_t, c_t),
    \label{eq:standard_vla}
\end{equation}
which unifies perception and action generation but does not explicitly model scene structure, dynamics, or intermediate reasoning states.

\paragraph{Structured latent formulation.}
To enable structured reasoning, we introduce multi-expert latent world representations
\(\mathcal{Z} = \{z_{\mathrm{sem}}, z_{\mathrm{geo}}, z_{\mathrm{dyn}}, z_{\mathrm{traj}}\}\),
encoding semantic interaction, geometric structure, dynamic evolution, and ego-trajectory priors. These form a Latent CoT bridging environment understanding and trajectory planning. Trajectory generation is then reformulated as
\begin{equation}
    p_\theta(\mathbf{A}_{t+1:t+T} \mid o_t, c_t, \mathcal{Z}).
    \label{eq:coworld_vla_formulation}
\end{equation}
In CoWorld-VLA, \(\mathcal{Z}\) is instantiated by the expert-token hidden states learned in Stage 2, where
\(z_{\mathrm{sem}}\), \(z_{\mathrm{geo}}\), \(z_{\mathrm{dyn}}\), and \(z_{\mathrm{traj}}\)
are represented by
\(H_{\mathrm{sem}}\), \(H_{\mathrm{geo}}\), \(H_{\mathrm{dyn}}\), and \(H_{\mathrm{traj}}\), respectively.

\subsection{Stage 1: Action-conditioned predictive world model}

We train an action-conditioned world model in the Wan latent space to predict future latent evolution from historical frames \(\mathbf{x}_h\) and ego intention, using future frames \(\mathbf{x}_f\) as targets.

\paragraph{Latent world modeling.}
Given a training video sequence \(\mathbf{x}=[\mathbf{x}_h,\mathbf{x}_f]\), we encode it using a frozen Wan VAE as
\begin{equation}
    \mathbf{z}
    =
    \mathcal{E}_{\mathrm{vae}}(\mathbf{x})
    =
    [\mathbf{z}_h,\mathbf{z}_f],
\end{equation}
where \(\mathbf{z}_h\) and \(\mathbf{z}_f\) denote historical and future latents, respectively. We apply flow matching only to the future segment. Given a noise level \(\sigma\in(0,1)\) and Gaussian noise \(\boldsymbol{\epsilon}\sim\mathcal{N}(0,\mathbf{I})\), the perturbed future latent and velocity target are defined as
\begin{equation}
    \tilde{\mathbf{z}}_{f,\sigma}
    =
    (1-\sigma)\mathbf{z}_f+\sigma\boldsymbol{\epsilon},
    \qquad
    \mathbf{v}_{\mathrm{target}}
    =
    \boldsymbol{\epsilon}-\mathbf{z}_f.
\end{equation}
The historical latent \(\mathbf{z}_h\) remains noise-free and serves as observed context.

\paragraph{Text-conditioned learning.}
To condition future scene evolution on ego intention and motion, we construct a structured prompt and encode it with the frozen Wan text encoder:
\begin{equation}
    \mathcal{P}
    =
    [\mathrm{Scene}]
    \oplus
    [\mathrm{Speed}]
    \oplus
    [\mathrm{Navigation}]
    \oplus
    [\mathrm{Trajectory}],
    \qquad
    \mathbf{c}
    =
    \mathcal{E}_{\mathrm{text}}(\mathcal{P}).
\end{equation}
The world model learns the conditional distribution \(p_{\theta}(\mathbf{z}_f \mid \mathbf{z}_h,\mathbf{c})\). With \(\tilde{\mathbf{z}}_{\sigma}=[\mathbf{z}_h,\tilde{\mathbf{z}}_{f,\sigma}]\), the flow matching loss is computed only on future-token outputs:
\begin{equation}
    \mathcal{L}_{\mathrm{flow}}
    =
    \mathbb{E}_{\mathbf{z}_h,\mathbf{z}_f,\boldsymbol{\epsilon},\sigma,\mathbf{c}}
    \left[
    \left\|
    \mathcal{F}_\theta
    \left(
    \tilde{\mathbf{z}}_{\sigma},
    \mathbf{c},
    \sigma
    \right)_f
    -
    \left(
    \boldsymbol{\epsilon}-\mathbf{z}_f
    \right)
    \right\|_2^2
    \right],
\end{equation}
where \((\cdot)_f\) denotes the future segment. This objective focuses learning on future dynamics while using historical observations as deterministic context.

\subsection{Stage 2: Multi-expert representation learning}

To provide VLM latent reasoning with sufficient physical and semantic priors, we introduce a multi-expert world representation learning framework. As shown in Stage 2 of Figure~\ref{fig:overview}, Qwen3-VL serves as the backbone, whose hidden states are aligned with semantic, geometric, visual-dynamic, and trajectory-level experts in a unified latent space.

\paragraph{Action representation generation.}
Given the current image observation \(o_t\) and the driving task prompt \(c_t\), the Qwen3-VL visual encoder \(V_{\mathrm{Qwen}}\) and tokenizer \(T_{\mathrm{Qwen}}\) produce image and text embeddings \(e_{\mathrm{img}}=V_{\mathrm{Qwen}}(o_t)\) and \(e_{\mathrm{txt}}=T_{\mathrm{Qwen}}(c_t)\), respectively.
In addition to the visual-textual input, we insert expert-specific action tokens to inject external physical priors into the VLM $\pi_\theta$ latent space:
\begin{equation}
\{H_{\mathrm{ctx}}, H_{\mathrm{sem}}, H_{\mathrm{geo}}, H_{\mathrm{dyn}}, H_{\mathrm{traj}}\}
=
\pi_\theta
\left(
e_{\mathrm{img}},
e_{\mathrm{txt}},
t_{\mathrm{sem}},
t_{\mathrm{geo}},
t_{\mathrm{dyn}},
t_{\mathrm{traj}}
\right),
\end{equation}
where \(H_{\mathrm{ctx}}\) denotes the contextual VLM hidden states; \(H_{\mathrm{sem}}\), \(H_{\mathrm{geo}}\), \(H_{\mathrm{dyn}}\), and \(H_{\mathrm{traj}}\) correspond to JEPA distillation, VGGT alignment, video model conditioning, and trajectory regression, respectively.

\paragraph{Multi-expert representation supervision.}
 We introduce three complementary expert branches to supervise VLM hidden states during training. The JEPA branch uses a frozen V-JEPA encoder to provide high-level semantic and predictive representations. The VGGT branch uses a frozen 3D foundation model to provide spatial and geometric priors. The Wan branch uses the pre-trained video generation world model to supervise visual-dynamic representations through future scene prediction.
For JEPA and VGGT, expert features are extracted from future observations \(o_{\mathrm{fut}}\) and pooled as supervision targets:
\begin{equation}
Z_{\mathrm{sem}}
=
\operatorname{Pool}
\left(
E_{\mathrm{sem}}(o_{\mathrm{fut}})
\right),
\quad
Z_{\mathrm{geo}}
=
\operatorname{Pool}
\left(
E_{\mathrm{geo}}(o_{\mathrm{fut}})
\right).
\end{equation}
The corresponding VLM action-token representations are adapted into expert feature spaces and optimized by alignment losses:
\begin{equation}
\mathcal{L}_{\mathrm{sem}}
=
\lambda_{\mathrm{l1}}
\operatorname{SmoothL1}
\left(
\hat{Z}_{\mathrm{sem}},
Z_{\mathrm{sem}}
\right)
+
\lambda_{\mathrm{cos}}
\left(
1
-
\cos
\left(
\hat{Z}_{\mathrm{sem}},
Z_{\mathrm{sem}}
\right)
\right),
\end{equation}
\begin{equation}
\mathcal{L}_{\mathrm{geo}}
=
\operatorname{MSE}
\left(
\hat{Z}_{\mathrm{geo}},
Z_{\mathrm{geo}}
\right),
\end{equation}
where \(\hat{Z}_{\mathrm{jepa}}\) and \(\hat{Z}_{\mathrm{geo}}\) are adapted VLM representations aligned with JEPA and VGGT features $Z_{\mathrm{jepa}}$ and $Z_{\mathrm{geo}}$.

For the Wan branch, \(H_{\mathrm{dyn}}\) serves as the conditioning signal for future scene generation, replacing the text-based condition adopted in Stage 1, and this branch is optimized by the flow matching objective:
\begin{equation}
\hat{o}_{\mathrm{fut}}
=
W_{\psi}
\left(
o_t,
H_{\mathrm{dyn}}
\right), \quad
\mathcal{L}_{\mathrm{dyn}}
=
\mathcal{L}_{\mathrm{flow}}
\left(
\hat{o}_{\mathrm{fut}},
o_{\mathrm{fut}};
o_t,
H_{\mathrm{dyn}}
\right).
\end{equation}

\paragraph{Trajectory prediction and joint optimization.}
Constrained by multi-expert supervision, the trajectory-aware representation \(H_{\mathrm{traj}}\) is used for planning-oriented regression. Specifically, the last token is fed into a lightweight MLP head to regress future $A$ trajectory waypoints, and the prediction is optimized with an MSE loss:
\begin{equation}
\hat{A}_{t+1:t+T}
=
\Phi_{\mathrm{traj}}
\left(
H_{\mathrm{traj}}[-1]
\right), \quad    
\mathcal{L}_{\mathrm{traj}}
=
\operatorname{MSE}
\left(
\hat{A}_{t+1:t+T},
A_{t+1:t+T}
\right), 
\end{equation}
where $\hat{A}_{t+1:t+T} \in \mathbb{R}^{T \times 3}$ denotes the predicted future trajectory, $\Phi_{\mathrm{traj}}: \mathbb{R}^{D} \rightarrow \mathbb{R}^{T \times 3}$ denotes the trajectory prediction head, and $D$ denotes the hidden-space dimension of the VLM.

The overall training objective is formulated as the weighted sum of all branches:
\begin{equation}
\mathcal{L}_{\mathrm{total}}
=
w_{\mathrm{dyn}}
\mathcal{L}_{\mathrm{dyn}}
+
w_{\mathrm{sem}}
\mathcal{L}_{\mathrm{sem}}
+
w_{\mathrm{geo}}
\mathcal{L}_{\mathrm{geo}}
+
w_{\mathrm{traj}}
\mathcal{L}_{\mathrm{traj}}, 
\end{equation}
where $w_{\mathrm{dyn}}$, $w_{\mathrm{sem}}$, $w_{\mathrm{geo}}$, and $w_{\mathrm{traj}}$ denote the loss weights of the Wan world model branch, JEPA representation distillation branch, VGGT geometric distillation branch, and trajectory prediction branch, respectively.

At inference time, CoWorld-VLA takes the current observation \(o_t\), the driving task prompt \(c_t\), and expert-specific action tokens as inputs, with no access to future information.

\subsection{Stage 3: Hierarchical multi-expert fusion}

Given the multi-expert tokens learned in Sec.~3.3, CoWorld-VLA generates continuous ego trajectories with a Hierarchical Multi-Expert Fusion (HMEF) planner. HMEF performs conditional denoising in the normalized action space, allowing expert tokens to directly guide trajectory generation.

HMEF takes scene tokens and expert action tokens as inputs. The scene tokens include scene context VLM tokens and current-frame JEPA and VGGT tokens extracted by the frozen encoders used in Stage 2, denoted as $H_{\mathrm{scene}}$. The expert action tokens are produced by the VLM:
\begin{equation}
\mathcal{H}_{\mathrm{act}}=\{H_{\mathrm{sem}}, H_{\mathrm{geo}}, H_{\mathrm{dyn}}, H_{\mathrm{traj}}\},
\end{equation}
corresponding to semantic, geometric, dynamic world-model, and trajectory-prior action cues.

\paragraph{Expert encoding.}
All token groups are first projected into a shared hidden space. To reduce the cost of denoising over long scene-token sequences, HMEF compresses $H_{\mathrm{scene}}$ into a fixed number of latent context tokens $C$ using learnable Perceiver queries~\cite{jaegle2021perceivergeneralperceptioniterative}. For each expert branch, an expert-specific bidirectional Transformer encodes the action tokens and aligns them to the planning horizon. Tokens associated with the same future step are projected and averaged to obtain per-step expert features $F_{e,t}$.

\paragraph{Conditional action fusion and denoising.}
The target trajectory $A=\{(x_t,y_t,\psi_t)\}_{t=1}^{T}$ is normalized to $A^{\mathrm{norm}}$. During training, HMEF samples $\tau\in(0,1)$ from a logit-normal schedule and constructs the noisy action:
\begin{equation}
A_\tau=(1-\tau)\epsilon+\tau A^{\mathrm{norm}},\quad \epsilon\sim\mathcal{N}(0,I).
\end{equation}
For each expert $e$ and future step $t$, HMEF forms an action token $R_{e,t}$ from the noisy action, timestep embedding $e_\tau$, ego-state information, and expert feature. The denoiser adopts a two-stream architecture with clean scene tokens $C$ and noisy expert-conditioned action tokens $R$. Each block applies joint self-attention between the two streams, followed by stream-specific feed-forward layers:
\begin{equation}
X_0=\mathcal{D}_{\theta}(C,R,e_\tau).
\end{equation}
The action-stream output is decoded into clean trajectory predictions $\hat A_e$ for all experts and supervised by the normalized ground truth:
\begin{equation}
\mathcal{L}_{\mathrm{diff}}
=
\frac{1}{N_e}\sum_{e=1}^{N_e}
\left\|\hat A_e-A^{\mathrm{norm}}\right\|_2^2 .
\end{equation}

\paragraph{Inference.}
HMEF learns expert fusion weights $\alpha=\operatorname{softmax}(w)$ and optimizes the fused prediction together with the expert denoising objective:
\begin{equation}
\bar A=\sum_{e=1}^{N_e}\alpha_e\hat A_e,\quad
\mathcal{L}_{\mathrm{act}}
=
\mathcal{L}_{\mathrm{diff}}
+
\lambda_{\mathrm{fusion}}\left\|\bar A-A^{\mathrm{norm}}\right\|_2^2 .
\end{equation}
At inference time, HMEF starts from Gaussian noise, iteratively denoises the expert trajectories, fuses them with $\alpha$, and denormalizes the result to obtain the executable ego plan.

\section{Experiments}
\subsection{Experimental settings}
\paragraph{Datasets.}
Following previous studies~\cite{1,2,3,4,5}, we train CoWorld-VLA on NuPlan~\cite{6} and NAVSIM v1~\cite{7}, evaluating future video generation and trajectory planning on NAVSIM v1. NuPlan contains approximately 1,200 hours of real-world driving data from four cities. NAVSIM v1, built upon OpenScene~\cite{8}, provides 120 hours of 2 Hz multi-view driving data for planning-oriented evaluation under challenging dynamic scenarios, with 1,192 training clips and 136 testing clips.

\paragraph{Metrics.}
For video generation, we report FVD~\cite{9} on NAVSIM v1. For trajectory planning, NAVSIM v1 adopts a non-reactive open-loop protocol with PDMS~\cite{7} as the primary metric, combining safety constraints and driving quality, including no at-fault collision (NC), drivable area compliance (DAC), ego progress (EP), time-to-collision (TTC), and comfort (C).

\paragraph{Implementation details.}
CoWorld-VLA consists of a video diffusion Transformer (Wan2.2-5B~\cite{11}), a VLM (Qwen3-VL-2B~\cite{12}), and an action expert network. We adopt a three-stage training strategy: video DiT pretraining on NuPlan for future video generation, VLM fine-tuning with multi-expert supervision on NAVSIM v1, and action expert training with the VLM frozen. During inference, 20 sampling steps are used for video generation and 10 for trajectory planning. $w_{\mathrm{dyn}}$=1.0, $w_{\mathrm{sem}}$=0.1, $w_{\mathrm{geo}}$=0.1, and $w_{\mathrm{traj}}$=1.0. More details are shown in the appendix.
\begin{table*}[t]
\centering
\setlength{\tabcolsep}{2.5pt}
\caption{Performance comparison on the NAVSIM v1 navtest under the benchmark planning protocol. PDMS and its sub-metrics evaluate the overall driving capability. Best and second-best results are highlighted in bold and underlined, respectively. C and L denote Camera and LiDAR, respectively. $^{\dagger}$ denotes results w/o reinforcement learning. $^{\P}$ denotes our method integrated with the ReCogDrive~\cite{2} action expert. }
\label{tab:navsim}
\begin{tabular}{@{}l c c c c c c c c c@{}}
\toprule
Method & Ref & Sensors & Frames & NC$\uparrow$ & DAC$\uparrow$ & TTC$\uparrow$ & Comf.$\uparrow$ & EP$\uparrow$ & \cellcolor{gray!15}PDMS$\uparrow$ \\
\midrule

\multicolumn{10}{l}{\textbf{End-to-End Methods}} \\

UniAD~\cite{74} & CVPR'23 & C & \textit{N} & 97.8 & 91.9 & 92.9 & \textbf{100} & 78.8 & 83.4 \\
Hydra-MDP~\cite{79} & arXiv'24 & C \& L & \textit{N} & 98.3 & 96.0 & 94.6 & \textbf{100} & 78.7 & 86.5 \\
DiffusionDrive~\cite{81} & CVPR'25 & C \& L & \textit{N} & 98.2 & 96.2 & 94.7 & \textbf{100} & 82.2 & 88.1 \\
TrajDiff~\cite{86} & arXiv'25 & C \& L & \textit{N} & 98.1 & \textbf{97.0} & 94.3 & \textbf{100} & 82.7 & 88.5 \\

\midrule
\multicolumn{10}{l}{\textbf{World Model Methods}} \\

LAW~\cite{82} & ICLR'25 & C & \textit{N} & 96.4 & 95.4 & 88.7 & \underline{99.9} & 81.7 & 84.6 \\
FSDrive~\cite{1} & NeurIPS'25 & C & \textit{N} & 98.2 & 93.8 & 93.3 & \underline{99.9} & 80.1 & 85.1 \\
Epona~\cite{5} & ICCV'25 & C & \textit{N} & 97.9 & 95.1 & 93.8 & \underline{99.9} & 80.4 & 86.2 \\
Resim~\cite{83} & NeurIPS'25 & C & \textit{N} & -- & -- & -- & -- & -- & 86.6 \\
PWM~\cite{72} & NeurIPS'25 & C & \textit{N} & 98.6 & 95.9 & 95.4 & \textbf{100} & 81.8 & 88.1 \\
WoTE~\cite{85} & ICCV'25 & C \& L &  \textit{N} & 98.5 & 96.8 & 94.9 & \underline{99.9} & 81.9 & 88.3 \\
ResWorld~\cite{78} & ICLR'26 & C \& L & \textit{N} & 98.9 & 96.5 & 95.6 & \textbf{100} & 83.1 & 89.0 \\
WorldDrive~\cite{87} & arXiv'26 & C & \textit{N} & 98.4 & 96.8 & 95.2 & \textbf{100} & 83.3 & 89.0 \\
DriveLaW~\cite{43} & CVPR'26 & C & \textit{N} & \underline{99.0} & 97.1 & \textbf{96.7} & \textbf{100} & 81.3 & 89.1 \\

\midrule
\multicolumn{10}{l}{\textbf{Vision-Language Model Methods}} \\

ReCogDrive$^{\dagger}$~\cite{2} & ICLR'26 & C & 1 &  98.1 & 94.7 & 94.2 & \textbf{100} & 80.9 & 86.5 \\
DriveVLA-W0~\cite{36} & ICLR'26 & C & \textit{N} & 98.4 & 95.3 & 95.4 & \textbf{100} & 80.9 & 87.2 \\
LaST-VLA$^{\dagger}$~\cite{luo2026lastvla} & arXiv'26 & C & 1 & 98.7 & 95.4 & 95.7 & \textbf{100} & 80.5 & 87.3 \\
SGDrive$^{\dagger}$~\cite{84} & CVPR'26 & C & \textit{N} & 98.6 & 95.1 & 95.4 & \textbf{100} & 81.2 & 87.4 \\
Uni-World VLA~\cite{4} & arXiv'26 & C & \textit{N} & 98.7 & 96.7 & 96.1 & \textbf{100} & 83.2 & \underline{89.4} \\

CoWorld-VLA$^{\P}$ & -- & C & 1 & 98.5 & \underline{96.9} & 95.4 & \textbf{100} & 83.2 & 89.1 \\
CoWorld-VLA (ours) & -- & C & 1 & \textbf{99.1} & \underline{97.0} & \underline{96.5} & \textbf{100} & \textbf{84.0} & \cellcolor{gray!15}\textbf{90.0} \\

\bottomrule
\end{tabular}
\end{table*}

\subsection{Main results}
\paragraph{Results on NAVSIM.}

Table~\ref{tab:navsim} presents planning results on NAVSIM v1 under its benchmark evaluation protocol. Under the single-frame front-camera-only setting, CoWorld-VLA achieves a PDMS of 90.0, remaining competitive with VLA-based planners such as SGDrive and Uni-World VLA, as well as world-model-based approaches including ResWorld and DriveLaW. These results support the effectiveness of world-model-guided latent representations for planning without multi-frame or LiDAR inputs.

For individual metrics, CoWorld-VLA achieves NC and EP scores of 99.1 and 84.0, respectively, indicating strong collision avoidance while maintaining forward progress. Its competitive DAC and TTC scores further suggest that the planned trajectories remain feasible and safety-aware. The results demonstrate the effectiveness of world-model-guided latent planning.

\paragraph{Results on NAVSIM v2 with EPDMS.}
As a complement to the NAVSIM v1 PDMS results, performance is additionally evaluated on the separate NAVSIM v2 navtest protocol using the extended PDMS metrics. 
Table~\ref{tab:epdms_navsim_v2} reports the corresponding comparison. 

\begin{table*}[t]
\centering
\setlength{\tabcolsep}{1.7pt}
\caption{Results on NAVSIM v2 navtest with extended PDMS metrics. EPDMS$^{*}$ is computed before the benchmark bug fix and EPDMS after it; ``--'' indicates unavailable results.}
\label{tab:epdms_navsim_v2}
\begin{tabular}{@{}l c c c c c c c c c c c@{}}
\toprule
Method & NC$\uparrow$ & DAC$\uparrow$ & DDC$\uparrow$ & TL$\uparrow$ & EP$\uparrow$ & TTC$\uparrow$ & LK$\uparrow$ & HC$\uparrow$ & EC$\uparrow$ & EPDMS$^{*}\uparrow$ & \cellcolor{gray!15}EPDMS$\uparrow$ \\
\midrule
\multicolumn{12}{l}{\textbf{End-to-End Methods}} \\
TransFuser & 96.9 & 89.9 & 97.8 & 99.7 & 87.1 & 95.4 & 92.7 & 98.3 & 87.2 & 76.7 & -- \\
DiffusionDrive & 98.2 & 95.9 & 99.4 & 99.8 & 87.5 & 97.3 & 96.8 & 98.3 & 87.7 & -- & 84.5 \\
DriveSuprim & 97.8 & 97.9 & 99.5 & 99.9 & 90.6 & 97.1 & 96.6 & 98.3 & 77.9 & 86.0 & -- \\
DiffusionDriveV2 & 97.7 & 96.6 & 99.2 & 99.8 & 88.9 & 97.2 & 96.0 & 97.8 & 91.0 & 85.5 & 87.5 \\
\midrule
\multicolumn{12}{l}{\textbf{World Model Methods}} \\
WoTE & 98.5 & 96.8 & 98.8 & 99.8 & 86.1 & 97.9 & 95.5 & 98.3 & 82.9 & -- & 87.7 \\
DreamerAD & 98.0 & 97.2 & 99.5 & 99.8 & 87.8 & 97.4 & 97.5 & 98.3 & 72.4 & -- & 87.7 \\
PWM & 98.8 & 95.9 & 99.4 & 99.9 & 86.4 & 98.4 & 97.6 & 98.3 & 85.3 & -- & 88.2 \\
DriveLaW & 98.7 & 96.9 & 99.6 & 99.8 & 87.5 & 98.3 & 97.6 & 98.4 & 77.4 & -- & 88.6 \\
Drive-JEPA (R34) & 98.8 & 97.4 & 99.0 & 99.8 & 83.5 & 98.0 & 96.2 & 98.1 & 85.6 & 85.4 & -- \\
Latent-WAM & 98.1 & 97.3 & 99.6 & 99.8 & 87.7 & 97.3 & 97.6 & 98.1 & 87.3 & -- & 89.3 \\
\midrule
\multicolumn{12}{l}{\textbf{Vision-Language Model Methods}} \\
WAM-Flow & 98.5 & 94.5 & 99.5 & 99.8 & 86.9 & 96.8 & 97.4 & 97.6 & 73.9 & 84.7 & -- \\
ReCogDrive$^\dagger$ & 98.3 & 95.2 & 98.3 & 99.8 & 87.1 & 97.5 & 96.6 & 99.5 & 86.5 & 83.6 & -- \\
DriveVLA-W0 & 98.5 & 99.1 & 98.0 & 99.7 & 86.4 & 98.1 & 93.2 & 97.9 & 58.9 & -- & 86.1 \\
SGDrive$^\dagger$ & 98.6 & 94.3 & 99.5 & 99.9 & 86.0 & 97.9 & 96.1 & 98.3 & 85.9 & -- & 86.2 \\
DriveWorld-VLA & 98.6 & 99.1 & 99.6 & 99.8 & 87.4 & 97.9 & 97.0 & 97.8 & 78.6 & -- & 86.8 \\
CoWorld-VLA (ours) & 99.1 & 97.0 & 99.6 & 99.9 & 87.8 & 98.5 & 97.7 & 98.2 & 86.2 & 86.2 & \cellcolor{gray!15}\textbf{90.0} \\
\bottomrule
\end{tabular}
\end{table*}

\paragraph{Evaluation of video generation results.}
In Stage 2, latent future world states are injected into the video DiT for future-aware video generation. Table~\ref{tab:fvd_compare} reports FVD results for CoWorld-VLA and prior driving video generation methods. CoWorld-VLA obtains an FVD of 32.7 on NAVSIM.

\begin{table}[!t]
\centering
\setlength{\tabcolsep}{1.8pt}
\caption{Reported FVD results across driving benchmarks. Lower FVD is better under matched data and evaluation protocols. Dataset and evaluation settings may differ across methods; cross-setting results are provided for contextual reference.}
\label{tab:fvd_compare}
\begin{tabular}{@{}lcccccc@{}}
\toprule
Method
& SVD\cite{89}
& GenAD\cite{90}
& DrivingGPT\cite{34}
& Epona\cite{5}
& DriveLaW\cite{43}
& CoWorld-VLA (ours) \\
\midrule
Dataset
& NAVSIM
& OpenDV
& NAVSIM
& NuPlan
& NuPlan
& NAVSIM \\
\cellcolor{gray!15}FVD$\downarrow$
& 227.5
& 184.0
& 142.6
& 61.3
& 55.6
& \cellcolor{gray!15}\textbf{32.7} \\
\bottomrule
\end{tabular}
\end{table}

\begin{figure*}[!t]
    \centering
    \includegraphics[width=\textwidth]{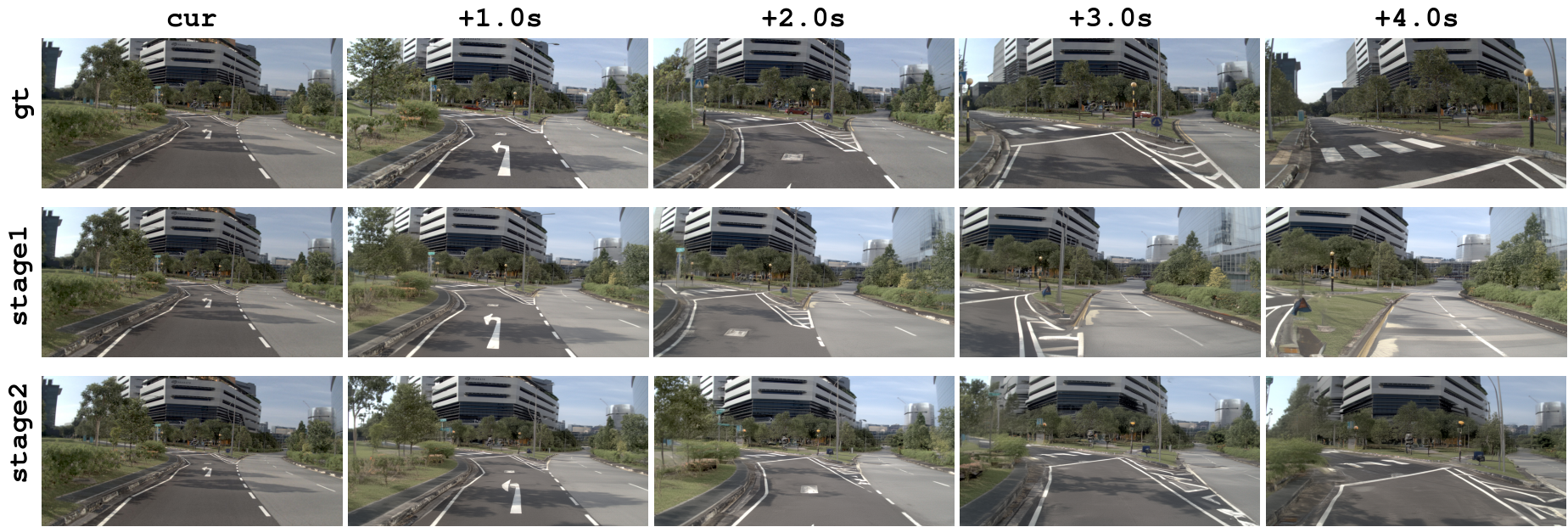}
    \caption{Qualitative comparison of future scene generation. Compared with Stage 1, Stage 2 better preserves the driving direction and lane-level scene evolution, producing future frames that are more consistent with the ground truth.}
    \label{fig:video generation}
\end{figure*}

\paragraph{Qualitative results of video generation.}
As shown in Figure~\ref{fig:video generation}, the Stage 1 model generates stable future scenes but may deviate from the ground-truth driving direction at intersections. With joint video DiT and VLM-based latent world modeling, the Stage 2 model better preserves turning behavior and road-layout evolution, producing predictions more consistent with the ground truth.

\begin{figure*}[!t]
    \centering
    \includegraphics[width=0.98\textwidth]{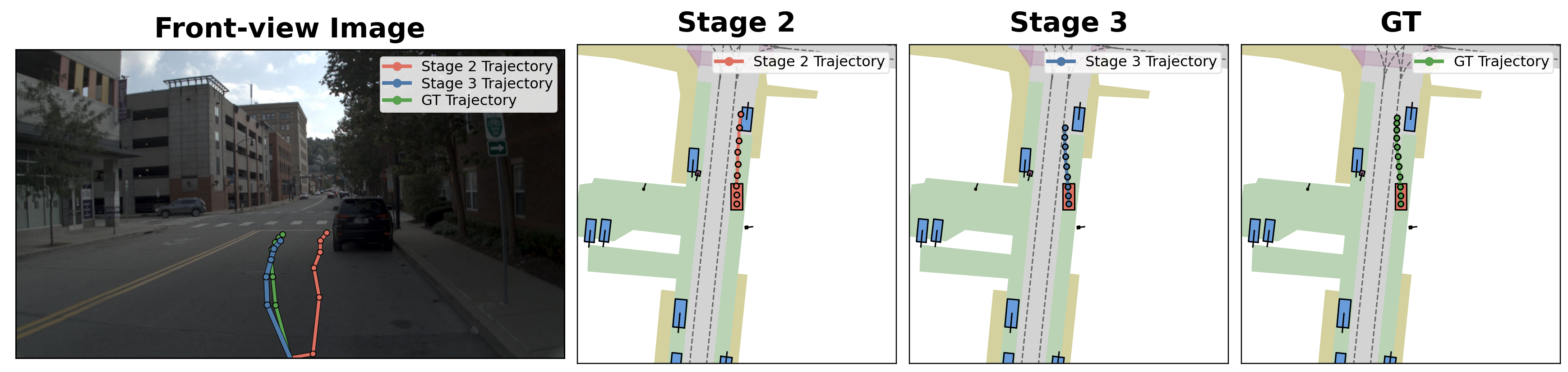}
    \vspace{0.25em}
    \includegraphics[width=0.98\textwidth]{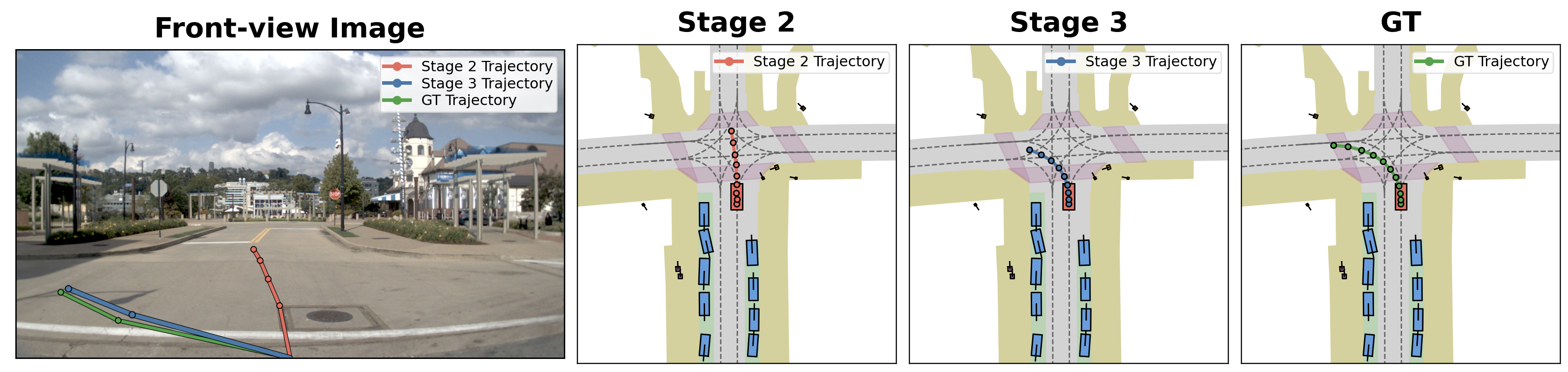}
    \caption{
    Qualitative comparison of trajectory planning across different training stages. Stage 2 predicts generally reasonable driving directions, but still shows deviations in lane keeping and turning scenarios. Stage 3 with HMEF generates trajectories that better align with the ground truth.
    }
    \label{fig:qualitative_traj_comparison}
\end{figure*}

\paragraph{Qualitative results of trajectory planning.}
As shown in Figure~\ref{fig:qualitative_traj_comparison}, Stage 2 can predict generally reasonable driving directions, but still shows deviations in lane keeping and turning scenarios. In comparison, Stage 3 with HMEF produces trajectories that are closer to the ground truth, especially in lateral position and turning tendency. These results show that HMEF further improves trajectory planning quality in complex driving scenes.

\FloatBarrier
\Needspace{18\baselineskip}
\subsection{Ablation study}
\paragraph{Ablation study on multi-expert design.}
\label{sec:ablation_multi_expert}

Table~\ref{tab:ablation_representation} reports the ablation results under the expert-token configurations listed from top to bottom. Using only the Ego Trajectory Token achieves a PDMS of 83.7. Adding the Geometric Structure Token alone improves PDMS to 85.1, while adding the Semantic Interaction Token alone gives 85.2. Starting from EgoT.+Sem., adding the Dynamic Evolution Token increases PDMS to 87.3; adding the Geometric Structure Token instead increases it to 87.7. Finally, using all four expert-token groups achieves 88.7, improving PDMS by 5.0 over EgoT. The two branches show that geometric, semantic, and dynamic supervision provide complementary planning cues rather than interchangeable information.

\begin{table}[H]
\centering
\setlength{\tabcolsep}{4pt}
\renewcommand{\arraystretch}{0.9}
\caption{Ablation study of latent world representations in Stage 2. EgoT., Geo., Sem., and Dyn. denote Ego Trajectory, Geometric Structure, Semantic Interaction, and Dynamic Evolution, respectively.}
\label{tab:ablation_representation}
\begin{tabular}{cccc|cccccc}
\toprule
EgoT. & Geo. & Sem. & Dyn.
& NC$\uparrow$ & DAC$\uparrow$ & TTC$\uparrow$
& Comf.$\uparrow$ & EP$\uparrow$ & \cellcolor{gray!15}PDMS$\uparrow$ \\
\midrule
\cmark &  &  &
& 97.7 & 92.7 & 92.7 & 100 & 78.5 & 83.7 \\
\cmark & \cmark &  &
& 97.7 & 93.9 & 92.7 & 100 & 80.3 & 85.1 \\
\cmark &  & \cmark &
& 98.1 & 93.7 & 93.8 & 100 & 79.0 & 85.2 \\
\cmark &  & \cmark & \cmark
& 98.3 & 95.4 & 95.1 & 100 & 81.1 & 87.3 \\
\cmark & \cmark & \cmark &
& 98.4 & 95.6 & 95.0 & 100 & 81.9 & 87.7 \\
\cmark & \cmark & \cmark & \cmark
& 98.4 & 96.5 & 95.3 & 100 & 82.3 & \cellcolor{gray!15}\textbf{88.7} \\
\bottomrule
\end{tabular}
\end{table}

\begin{table}[H]
\centering
\setlength{\tabcolsep}{1pt}
\renewcommand{\arraystretch}{0.9}
\caption{Ablation study of representation learning and planning modules on NAVSIM v1. Full Experts denotes the joint use of EgoT., Geo., Sem., and Dyn.; AE denotes the ReCogDrive~\cite{2} action expert.}
\label{tab:ablation_ae}
\begin{tabular*}{\textwidth}{@{\extracolsep{\fill}}cc|ccc|cccccc@{}}
\toprule
\multicolumn{2}{c|}{Representation}
& \multicolumn{3}{c|}{Planner}
& \multirow{2}{*}{NC$\uparrow$}
& \multirow{2}{*}{DAC$\uparrow$}
& \multirow{2}{*}{TTC$\uparrow$}
& \multirow{2}{*}{Comf.$\uparrow$}
& \multirow{2}{*}{EP$\uparrow$}
& \cellcolor{gray!15}\multirow{2}{*}{PDMS$\uparrow$} \\
\cmidrule(lr){1-2}\cmidrule(lr){3-5}
{\footnotesize EgoT. only} &
{\footnotesize Full Experts}
& {\footnotesize VLM only} &
{\footnotesize VLM+AE} &
{\footnotesize VLM+HMEF}
& & & & & & \\
\midrule
\cmark &  & \cmark &  &
& 97.7 & 92.7 & 92.7 & 100 & 78.5 & 83.7 \\
 & \cmark & \cmark &  &
& 98.4 & 96.5 & 95.3 & 100 & 82.3 & 88.7 \\
 & \cmark &  & \cmark &
& 98.5 & 96.9 & 95.4 & 100 & 83.2 & 89.1 \\
\cmark &  &  &  & \cmark
& 98.3 & 96.8 & 95.0 & 100 & 83.2 & 88.9 \\
 & \cmark &  &  & \cmark
& 99.1 & 97.0 & 96.5 & 100 & 84.0 & \cellcolor{gray!15}\textbf{90.0} \\
\bottomrule
\end{tabular*}
\end{table}

\paragraph{Ablation study on representation and planning modules.}
\label{sec:ablation_planner}

Table~\ref{tab:ablation_ae} jointly evaluates representation learning and trajectory planning. With VLM only fixed, Full Experts improves PDMS from 83.7 with EgoT only to 88.7. With Full Experts fixed, VLM+AE and VLM+HMEF achieve PDMS scores of 89.1 and 90.0, improving PDMS by 0.4 and 1.3 over VLM only, respectively. With VLM+HMEF fixed, EgoT only achieves 88.9 compared with 90.0 for Full Experts, indicating complementary benefits from the planner and multi-expert representations.

\paragraph{Denoising step ablation.}
\label{sec:ablation_denoising}

As shown in Table~\ref{tab:ablation_step}, we ablate the number of denoising steps in the fusion diffusion planner. The results suggest that 10 denoising steps provide the best balance between accuracy and efficiency.

\begin{table}[H]
\centering
\setlength{\tabcolsep}{4pt}
\caption{Effect of diffusion denoising steps.}
\label{tab:ablation_step}
\begin{tabular}{c|cccccc}
\toprule
& NC$\uparrow$ & DAC$\uparrow$ & TTC$\uparrow$
& Comf.$\uparrow$ & EP$\uparrow$ & \cellcolor{gray!15}PDMS$\uparrow$ \\
\midrule
Step = 5
& 99.0 & 96.8 & 95.9 & 100 & 83.6 & 89.5 \\
Step = 10
& 99.1 & 97.0 & 96.5 & 100 & 84.0 & \cellcolor{gray!15}\textbf{90.0} \\
Step = 20
& 99.1 & 96.7 & 96.4 & 100 & 83.5 & 89.7 \\
\bottomrule
\end{tabular}
\end{table}

\section{Conclusions}

We propose CoWorld-VLA, a multi-expert world reasoning framework. By encoding semantic, geometric, dynamic, and trajectory priors into expert tokens, it forms a planning-oriented Latent CoT. A hierarchical fusion planner then integrates these tokens to generate continuous trajectories. Experiments demonstrate effective future-scene modeling and strong planning performance in complex driving scenarios.

\textbf{Limitations}.

The primary limitation of our framework is the substantial computational overhead incurred during multi-stage training. This cost is concentrated in teacher-based pretraining and VLM representation learning; the Wan model is not used in Stage 3 or during planning inference, while the Stage-2 VLM and the V-JEPA/VGGT encoders are frozen during Stage 3. The current model is also limited to single-image inputs. Future work will investigate caching teacher features, lightweight dynamics teachers, parameter-efficient fine-tuning such as LoRA, and distillation or removal of online expert encoders, while extending the framework to multi-camera settings.

\bibliographystyle{unsrt} 
\bibliography{ref}


\appendix
\numberwithin{equation}{section} 

\section{Technical appendices and supplementary material}
\subsection{More implementation details}
\subsubsection{Additional details of stage 1 action-conditioned predictive world model}
This section provides additional details on the conditioning design of the Stage 1 predictive world model. In this stage, the word ``action-conditioned'' refers to text-form conditions that describe ego intention and motion, rather than low-level continuous control commands. These conditions are encoded by the frozen UMT5 text encoder from Wan and used to guide future scene prediction.

\paragraph{Condition prompt construction.}
As defined in Sec.~3.1, the condition prompt \(\mathcal{P}\) is composed of four components. Here, we detail how each component is serialized into natural language. \([\mathrm{Scene}]\) is a static scene prompt from the configuration, \([\mathrm{Speed}]\) describes the ego speed with a coarse natural-language phrase, \([\mathrm{Navigation}]\) provides the high-level driving command when available, and \([\mathrm{Trajectory}]\) serializes ego waypoints in the local ego-centric coordinate system. The current ego position is explicitly anchored at the origin to establish a clear coordinate frame for the text encoder.
This prompt is then encoded by the frozen UMT5 encoder and used as the conditioning signal for the predictive world model.

\paragraph{Optional fields and fallback behavior.}
The navigation command is optional. When it is available, it is converted into one of four textual commands: \texttt{turn left}, \texttt{go straight}, \texttt{turn right}, and \texttt{unknown}. When the navigation command is absent, the corresponding command phrase is omitted from the prompt. Similarly, the trajectory branch is used only when a valid future trajectory is provided. If no valid future trajectory is available, the condition encoder falls back to the configured scene prompt without adding the speed, command, or trajectory clauses.

\paragraph{Speed and trajectory representation.}
The ego speed is represented by a coarse phrase rather than a raw numerical value. Specifically, the speed is mapped into phrases such as \texttt{nearly stopped}, \texttt{driving slowly}, \texttt{driving at moderate speed}, and \texttt{driving at high speed}. When an explicit ego speed is unavailable, it can be estimated from ego-state information or approximated from the future trajectory displacement. The trajectory condition is represented as a sequence of future \((x,y)\) waypoints with fixed decimal precision. We omit heading values in the text prompt, since the polyline shape together with the navigation command already provides sufficient directional information.

\subsubsection{Additional details of stage 2 multi-expert representation learning}

This section provides additional details on the multi-expert representation learning stage. The main formulation is described in Sec.~3.3. Here, we focus on how expert tokens are organized inside the VLM and how different supervision branches shape planning-oriented latent representations.

\paragraph{Expert token organization.}

In Stage 2, we insert four groups of learnable expert tokens into the VLM input sequence, corresponding to semantic interaction, geometric structure, dynamic evolution, and ego trajectory. After the image tokens, text tokens, and expert tokens are processed by the VLM, the hidden states at the corresponding expert-token positions are extracted as \(H_{\mathrm{sem}}\), \(H_{\mathrm{geo}}\), \(H_{\mathrm{dyn}}\), and \(H_{\mathrm{traj}}\). These hidden states are not treated as textual outputs. Instead, they serve as continuous latent representations that form a planning-oriented Latent CoT.

\paragraph{Token-level expert alignment.}
For the JEPA and VGGT branches, the frozen expert models produce dense visual features rather than a single global vector. We therefore apply pooling to convert expert outputs into compact token sequences. The number of pooled expert tokens is kept consistent with the corresponding action-token group, so that each VLM expert token can be aligned with one target expert feature. Since the VLM hidden dimension and the expert feature dimension are generally different, lightweight projection modules are used to map \(H_{\mathrm{sem}}\) and \(H_{\mathrm{geo}}\) into the corresponding expert feature spaces before applying the alignment losses.

\paragraph{Semantic and geometric supervision.}
The JEPA branch provides high-level semantic supervision from future observations, encouraging the semantic interaction token to encode object-level context, scene semantics, and interaction-related information. The VGGT branch provides geometric supervision, encouraging the geometric structure tokens to encode road layout, spatial configuration, and 3D structural cues. These two branches complement each other: JEPA offers abstract semantic and temporal understanding, while VGGT provides explicit spatial grounding.

\paragraph{World-model supervision through action-token conditioning.}
For the dynamic evolution branch, the VLM-generated world-model tokens \(H_{\mathrm{dyn}}\) are used as conditional latent variables for the Wan world model. The VLM itself does not directly decode future images. Instead, future scene generation is performed by the Wan world model conditioned on \(H_{\mathrm{dyn}}\), and the flow-matching objective provides supervision to the corresponding action tokens. This design allows the dynamic tokens to learn future motion trends and temporal consistency without requiring the VLM backbone to act as a pixel-level generator.

\paragraph{Trajectory-token supervision.}
The trajectory tokens are supervised by a lightweight trajectory regression head. This branch encourages \(H_{\mathrm{traj}}\) to encode behavior-oriented information that is directly related to future ego motion. In Stage 2, this trajectory head is used to shape the latent action representation rather than serve as the final planner. The final trajectory generation is performed in Stage 3 by the hierarchical diffusion planner, which further integrates multi-expert representations.

\paragraph{Role of Stage 2.}
Overall, Stage 2 transforms heterogeneous expert supervision into structured latent states inside the VLM. Instead of relying only on language supervision, the VLM receives complementary constraints from semantic representation learning, geometric structure alignment, future scene generation, and trajectory regression. These expert tokens are then reused by the downstream action expert as planning-oriented latent conditions.

\subsubsection{Additional details of hierarchical multi-expert fusion}
This section provides additional implementation details of the Hierarchical Multi-Expert Fusion (HMEF) planner. The main formulation of HMEF is described in Sec.~3.4. Here, we focus on scene compression, per-step expert feature extraction, trajectory normalization, and fusion-weight training.

\paragraph{Scene context compression.}
The scene context tokens can be long and variable in length. To reduce the computational cost of the denoiser, each enabled scene-context source is compressed by its own Perceiver-style compressor. Specifically, non-action VLM tokens, current-frame JEPA context tokens, and current-frame VGGT context tokens are separately compressed into fixed-length latent tokens when enabled. These compressed tokens are then concatenated along the sequence dimension and used as the clean scene stream in the subsequent joint denoising process.

\paragraph{Per-step expert feature extraction.}
Each expert action-token group is first projected to the HMEF hidden dimension and processed by an expert-specific bidirectional Transformer. To align expert tokens with the planning horizon, we organize the output tokens by future timestep. If multiple tokens correspond to the same future step, their projected features are averaged to obtain one per-step expert feature. This yields a sequence of expert conditions aligned with the future waypoints, allowing each denoising step to receive timestep-specific semantic, geometric, dynamic, or trajectory-prior guidance.

\paragraph{Historical and ego-state conditioning.}
In addition to the expert tokens, HMEF also utilizes low-level ego information. The historical ego trajectory and the current ego status are separately encoded by lightweight MLPs and then fused into a single conditioning vector. This fused ego condition is expanded along the planning horizon and combined with the noisy action token and the corresponding expert feature before being passed into the denoiser. This design allows the planner to jointly leverage high-level abstract priors and low-level vehicle-state telemetry.

\paragraph{Trajectory normalization.}
The conditional denoising process is performed in a normalized action space. Specifically, the future trajectory \((x,y,\psi)\) is normalized to \([-1,1]\) using empirical coordinate ranges calculated from the NAVSIM dataset. During inference, the predicted trajectory is denormalized back to the original coordinate space before evaluation. This normalization stabilizes the training dynamics and keeps the coordinate dimensions on comparable mathematical scales.

\paragraph{Fusion-weight optimization.}
HMEF predicts one trajectory for each active expert branch and learns global scalar fusion weights to combine them. During the calculation of the fusion loss, the individual expert trajectories are detached from the computation graph before weighted averaging. Consequently, the fusion objective updates only the fusion weights rather than propagating gradients back into the individual expert denoising branches, while the expert branches are still trained by the denoising loss. This helps stabilize training by separating expert-specific trajectory generation from global expert-importance learning.

\subsubsection{Training details.}
CoWorld-VLA comprises three components: a video diffusion Transformer (Wan2.2-5B), a VLM (Qwen3-VL-2B), and an action expert network. We adopt a three-stage training strategy. First, the video DiT is pretrained on 8 Hz NuPlan videos for future video generation using 48k training steps, a batch size of 192, and a cosine learning rate schedule with warmup. Second, the VLM is fine-tuned on NAVSIM v1 with multi-expert supervision for 40k steps, using learning rates of \(2\times10^{-5}\) for the VLM, \(1\times10^{-4}\) for the JEPA adaptor, and \(1\times10^{-5}\) for newly introduced modules. Third, the action expert network is trained for 60k steps with the fine-tuned VLM frozen, using a batch size of 256 and a learning rate of \(2\times10^{-5}\). Stage 1 training is conducted on 64 NVIDIA A800 GPUs and requires approximately 74 hours. Stage 2 uses 32 NVIDIA A800 GPUs with approximately 70 training hours, while Stage 3 is trained on 16 NVIDIA A800 GPUs for approximately 20 hours.

\subsection{More experimental results}
\subsubsection{Inference-seed stability}
Holding the trained Stage 3 weights fixed, we evaluate six inference-time random seeds. Table~\ref{tab:seed_stability} reports means of 89.980 (standard deviation 0.013) for PDMS and 89.985 (standard deviation 0.014) for EPDMS. These results indicate low sensitivity to the evaluated inference seeds. They do not replace variance estimates over independently trained models, which were not conducted because of the high training cost.

\begin{table}[t]
\centering
\setlength{\tabcolsep}{5pt}
\caption{Performance stability across inference-time random seeds.}
\label{tab:seed_stability}
\begin{tabular}{lcccccc}
\toprule
Metric & Seed 42 & Seed 1 & Seed 2 & Seed 3 & Seed 4 & Seed 5 \\
\midrule
PDMS & 89.964 & 89.975 & 89.986 & 89.971 & 89.982 & 90.000 \\
EPDMS & 89.979 & 89.989 & 89.961 & 89.983 & 90.000 & 89.997 \\
\bottomrule
\end{tabular}
\end{table}

\subsubsection{Learnable Expert Weight Analysis}
The learnable expert weights are initialized uniformly at 0.25. After convergence, the weights evolve to approximately 0.35 for dynamic evolution expert, 0.19 for semantic interaction expert, 0.15 for geometric structure expert, and 0.31 for trajectory expert, indicating that the model automatically assigns higher importance to dynamic and trajectory-related representations during planning.
\subsubsection{Additional qualitative results on future video generation}
\begin{figure*}[!t]
    \centering
    \subfloat[Left Turn Navigation at a Forked Intersection]{
        \includegraphics[width=0.98\textwidth]{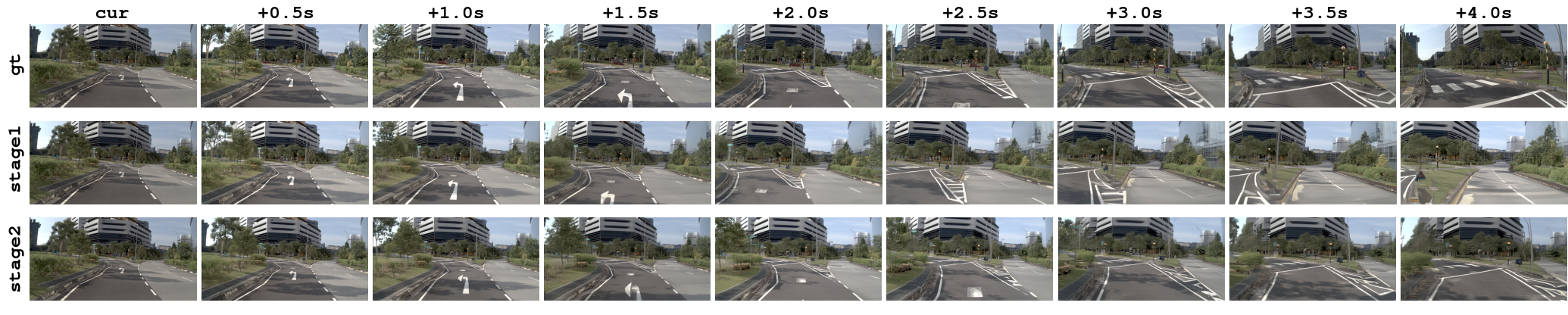}
        \label{fig:video_case_intersection}
    }

    \subfloat[Straight Cruising in a Multi-Lane Urban Street]{
        \includegraphics[width=0.98\textwidth]{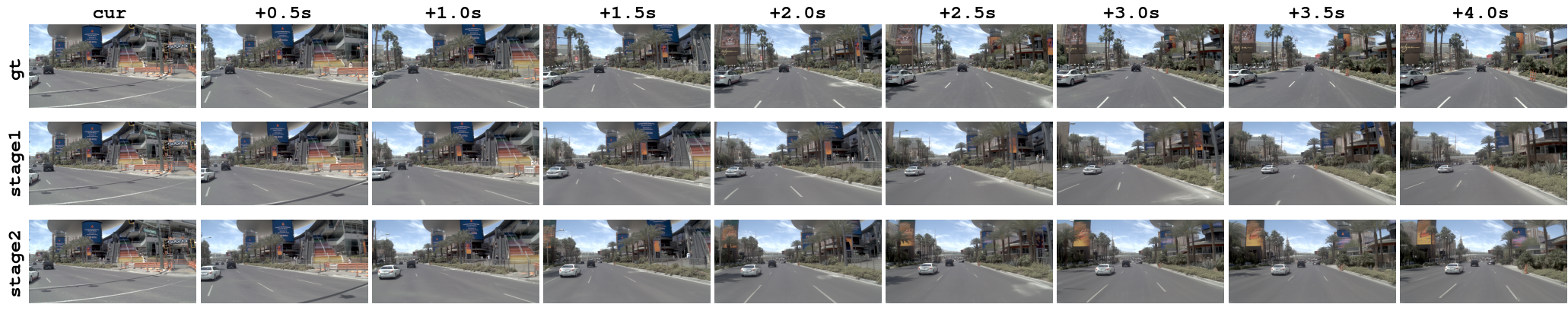}
        \label{fig:video_case_downtown}
    }

    \subfloat[Close-Proximity Car-Following in Dense Traffic]{
        \includegraphics[width=0.98\textwidth]{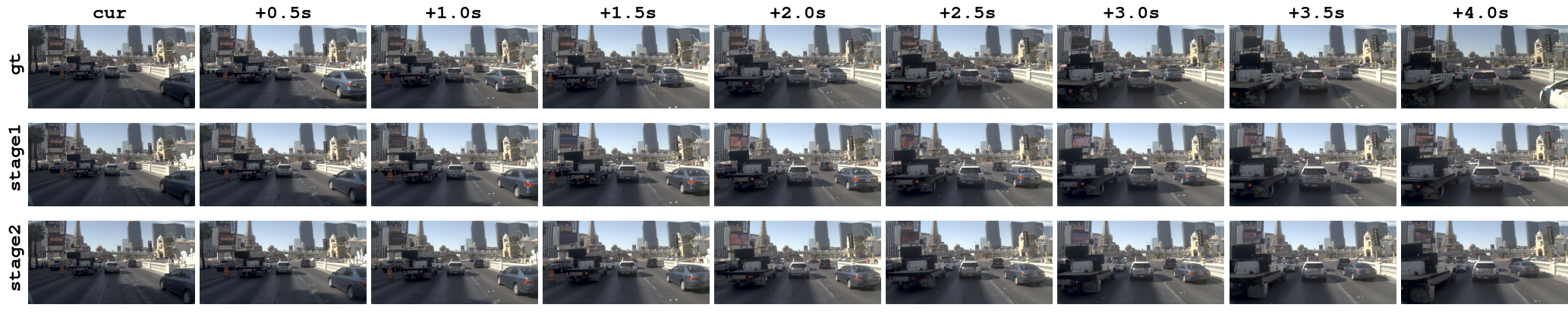}
        \label{fig:video_case_dense}
    }

    \caption{
    Additional qualitative results of future video generation under different driving scenarios.Each case compares the ground truth, Stage 1 prediction, and Stage 2 prediction over the future horizon.
    }
    \label{fig:video_cases}
\end{figure*}

We provide additional qualitative comparisons of future video generation in Figure~\ref{fig:video_cases}, covering three representative scenarios: left-turn at a forked intersection, straight cruising on a multi-lane urban road, and close-proximity car-following in dense traffic. For each scenario, we compare the ground truth (GT) with Stage 1 and Stage 2 predictions.

In the forked-intersection scenario (Figure~\ref{fig:video_cases}(a)), Stage 1 predicts a straight-driving future instead of the intended left turn. Stage 2 preserves the left-turn trajectory and aligns well with the GT, demonstrating the effect of multi-expert Latent CoT in constraining behaviorally relevant predictions.

For straight cruising (Figure~\ref{fig:video_cases}(b)), Stage 1 gradually drifts toward the adjacent lane, while Stage 2 maintains lane alignment and stable forward progression. This highlights the benefit of geometric structure supervision for road-layout consistency.

In the close-proximity car-following scenario (Figure~\ref{fig:video_cases}(c)), both Stage 1 and Stage 2 capture general vehicle motion, but Stage 2 better preserves relative spacing and motion continuity, showing the advantage of temporal dynamics supervision in dense traffic.

Overall, Stage 2 multi-expert training improves temporal consistency, preserves lane-level structure, and produces predictions that more faithfully follow ego intentions compared with Stage 1.

\begin{figure*}[!t]
    \centering
    \includegraphics[width=0.98\textwidth]{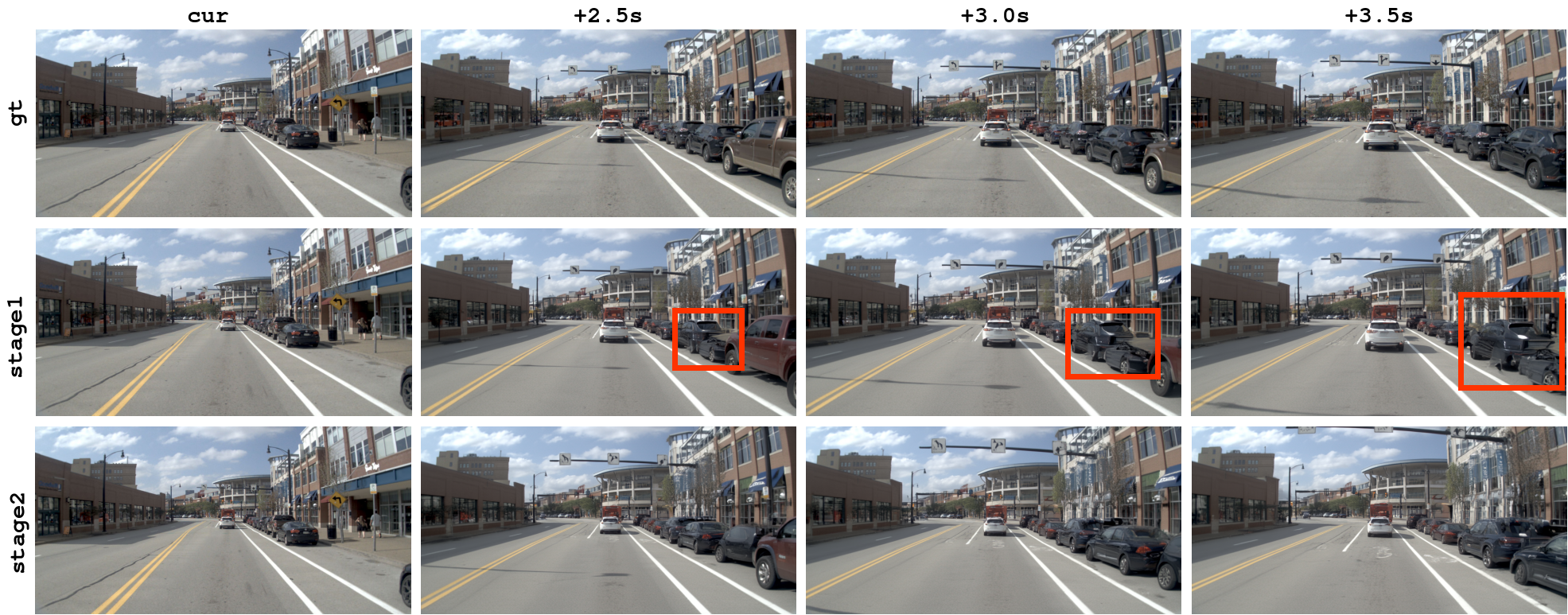}
    \caption{
    Local fidelity comparison in future video generation. The red boxes highlight roadside vehicles that become blurred and distorted in the Stage 1 prediction, while Stage 2 preserves clearer object boundaries and more stable local structure.
    }
    \label{fig:video_case_blur}
\end{figure*}

In addition, Figure~\ref{fig:video_case_blur} highlights a local fidelity comparison in a downtown driving scene. The red boxes mark parked vehicles on the right side of the road. While Stage 1 produces a plausible global road layout, the highlighted region becomes blurred and distorted in later frames, showing reduced stability in local object appearance. Stage 2 preserves clearer vehicle boundaries and more consistent roadside structure throughout the generated sequence. This case suggests that Stage 2 multi-expert supervision improves not only high-level driving direction and trajectory alignment, but also fine-grained visual consistency in future scene generation.

\subsubsection{Additional qualitative results on trajectory planning}

\begin{figure*}[!t]
    \centering
    \subfloat[Lane-Keeping Cruising Scenario]{
        \includegraphics[width=0.98\textwidth]{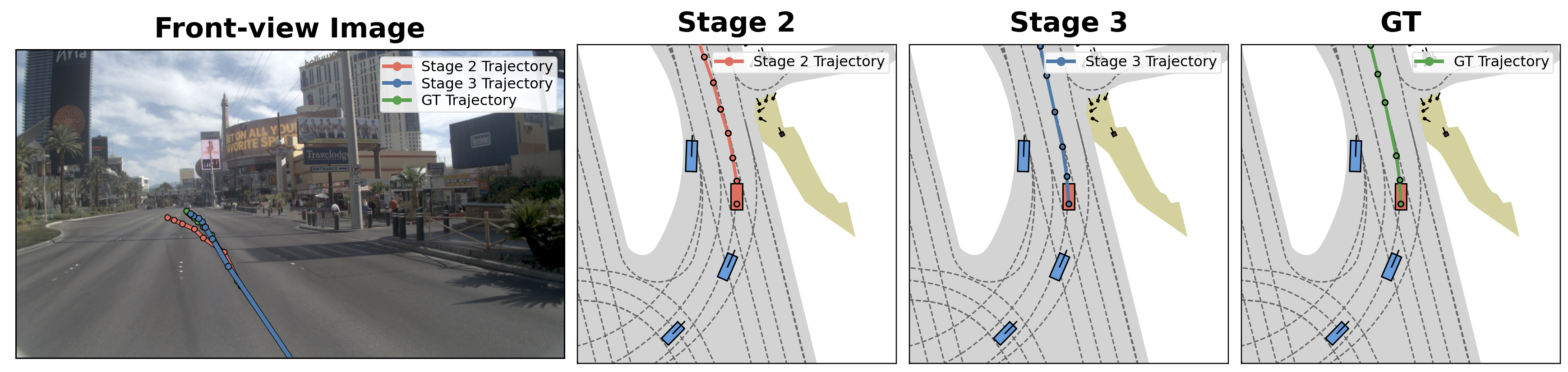}
        \label{fig:traj_case_straight}
    }

    \subfloat[Left-Turn Navigation at an Intersection ]{
        \includegraphics[width=0.98\textwidth]{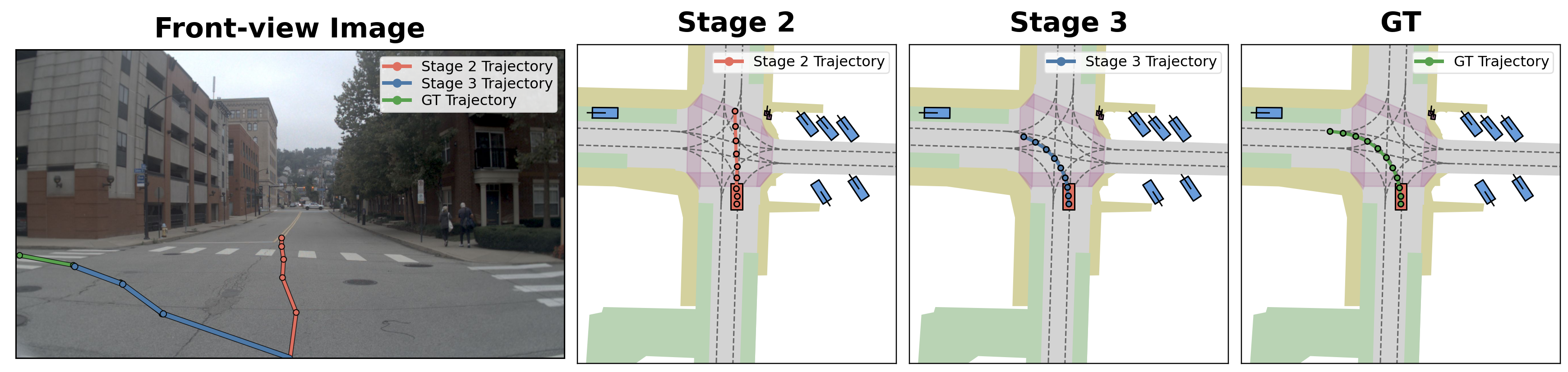}
        \label{fig:traj_case_left_turn}
    }

    \subfloat[Active Overtaking Maneuver in Dynamic Traffic]{
        \includegraphics[width=0.98\textwidth]{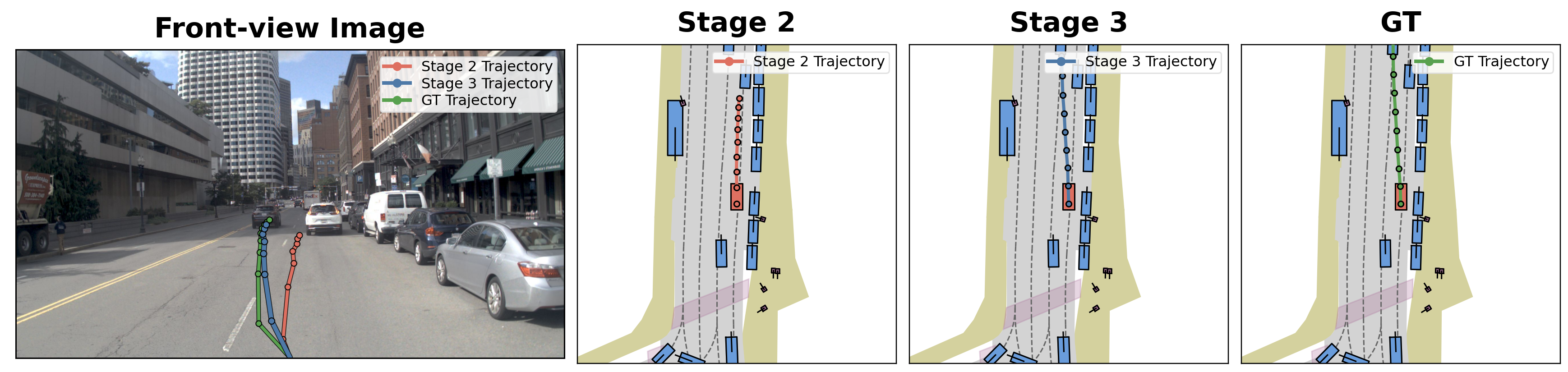}
        \label{fig:traj_case_overtake}
    }

    \caption{
    Qualitative comparison of trajectory planning across three representative driving scenarios. Each case illustrates the planned trajectories from the Stage 2 regression baseline and the Stage 3 HMEF planner, compared against the Ground Truth (GT) over the future horizon. The multi-expert fusion in Stage 3 significantly improves geometric adherence and dynamic interaction.
    }
    \label{fig:traj_case}
\end{figure*}

We provide qualitative comparisons of Stage 2 and Stage 3 trajectories across three representative driving scenarios in Figure~\ref{fig:traj_case}. Overall, Stage 2 captures coarse driving intentions, while Stage 3 produces more accurate and stable trajectories through hierarchical multi-expert fusion.

As shown in Figure~\ref{fig:traj_case}(a), the lane-keeping cruising scenario requires stable forward motion within the current lane. Stage 2 predicts the general driving direction but exhibits noticeable lateral drift over the planning horizon. In contrast, Stage 3 generates a more centered trajectory that closely follows the GT, indicating improved lane-level consistency and long-horizon stability.

Figure~\ref{fig:traj_case}(b) shows an intersection left-turn scenario, where planning requires both navigation-intention understanding and road-topology awareness. Stage 2 deviates from the desired turning behavior, while Stage 3 better follows the intersection layout and produces a smooth left-turn trajectory closer to the GT. This suggests that HMEF more effectively translates latent reasoning states into executable turning actions.

Figure~\ref{fig:traj_case}(c) presents a detour maneuver around a leading vehicle. Stage 2 captures part of the forward motion but fails to match the correct bypass trajectory. Stage 3 better preserves the required lateral offset and trajectory curvature, demonstrating stronger vehicle-aware planning ability.

Overall, these results show that Stage 3 consistently improves over Stage 2 in lane keeping, intersection turning, and detour planning. By fusing heterogeneous expert priors during diffusion-based action generation, HMEF better converts multi-expert Latent CoT representations into actionable and behavior-consistent ego trajectories.

\end{document}